\documentclass{article} 
\usepackage[preprint]{colm2026_conference}

\usepackage{microtype}
\usepackage{hyperref}
\usepackage{url}
\usepackage{booktabs}
\usepackage{amsmath}
\usepackage{amssymb}
\usepackage{multirow}
\usepackage{bbm}
\usepackage{graphicx}
\usepackage[table]{xcolor}
\usepackage{subcaption}
\usepackage{tcolorbox}
\usepackage{colortbl}
\usepackage{arydshln}
\usepackage{fancyvrb}
\usepackage{verbatim}
\usepackage{soul}
\usepackage{algorithm}
\usepackage{algpseudocode}

\usepackage{lineno}
\usepackage{fontawesome5}

\definecolor{darkblue}{rgb}{0, 0, 0.5}
\hypersetup{colorlinks=true, citecolor=darkblue, linkcolor=darkblue, urlcolor=darkblue}

\definecolor{hlyellow}{rgb}{1.0,0.95,0.6}
\definecolor{hlgreen}{rgb}{0.75,0.95,0.75}

\definecolor{promptbg}{rgb}{0.950,0.950,0.950}
\definecolor{promptframe}{rgb}{0.600,0.600,0.600}

\newtcolorbox{promptbox}[1][]{
    colback=promptbg,
    colframe=promptframe,
    boxrule=1pt,
    arc=3pt,
    boxsep=5pt,
    left=10pt,right=10pt,top=10pt,bottom=10pt,
    fonttitle=\bfseries,
    fontupper=\ttfamily\small,
    title={#1}
}

\title{Unlocking Prompt Infilling Capability for Diffusion Language Models}

\author{Yoshinari Fujinuma\thanks{Correspondence to \texttt{fujinumay@gmail.com}} \\
  Patronus AI \\
\And
  Keisuke Sakaguchi \\
  Tohoku University, RIKEN \\
}

\definecolor{promptcolor}{RGB}{66, 135, 245}
\definecolor{completioncolor}{RGB}{245, 135, 66}
\definecolor{blockcolor}{RGB}{200, 200, 200}

\begin{document}

\ifcolmsubmission
\linenumbers
\fi

\maketitle

\begin{abstract}
Masked diffusion language models (dLMs) generate text through bidirectional denoising, yet this capability remains locked for infilling prompts. This limitation is an artifact of the current supervised finetuning (SFT) convention of applying response-only masking.
To unlock this capability, we extend full-sequence masking during SFT, where both prompts and responses are masked jointly.
Once unlocked, the model infills masked portions of a prompt template conditioned on few-shot examples. We show that such model-infilled prompts match or surpass manually designed templates, transfer effectively across models, and are complementary to existing prompt optimization methods. Our results suggest that training practices, not architectural limitations, are the primary bottleneck preventing masked diffusion language models from infilling effective prompts.
\end{abstract}

\begin{center}
\href{https://github.com/akkikiki/dlm_prompt_infilling}{\faGithub\ Code}\quad\href{https://huggingface.co/collections/akkikiki/dlm-prompt-infilling}{\raisebox{-0.15em}{\includegraphics[height=1em]{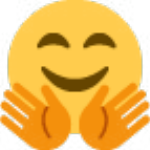}}~Models}
\end{center}

\section{Introduction}

Masked diffusion language models (dLMs)~\citep{lou2024sedd,sahoo2024mdlm,nie2025largelanguagediffusionmodels,ye2025dream7bdiffusionlarge} generate text through iterative bidirectional denoising, progressively unmasking tokens across the entire sequence. Unlike autoregressive models that generate strictly left-to-right, dLMs can fill in any masked position conditioned on all surrounding context. This \emph{infilling} capability is a defining advantage of dLMs.

Yet current dLMs cannot fully exploit this advantage. Models such as LLaDA~\citep{nie2025largelanguagediffusionmodels} and Dream~\citep{ye2025dream7bdiffusionlarge} are supervised-fine-tuned with response-only masking: prompts remain fully visible while only responses are masked and denoised. As a result, these models are trained to denoise responses given clean prompts but never trained to denoise prompts. This is not an architectural limitation but a training one: the bidirectional architecture already supports prompt-side denoising, yet response-only masking during SFT withholds the necessary training signal.

\begin{figure*}[t]
\centering
\includegraphics[width=\textwidth]{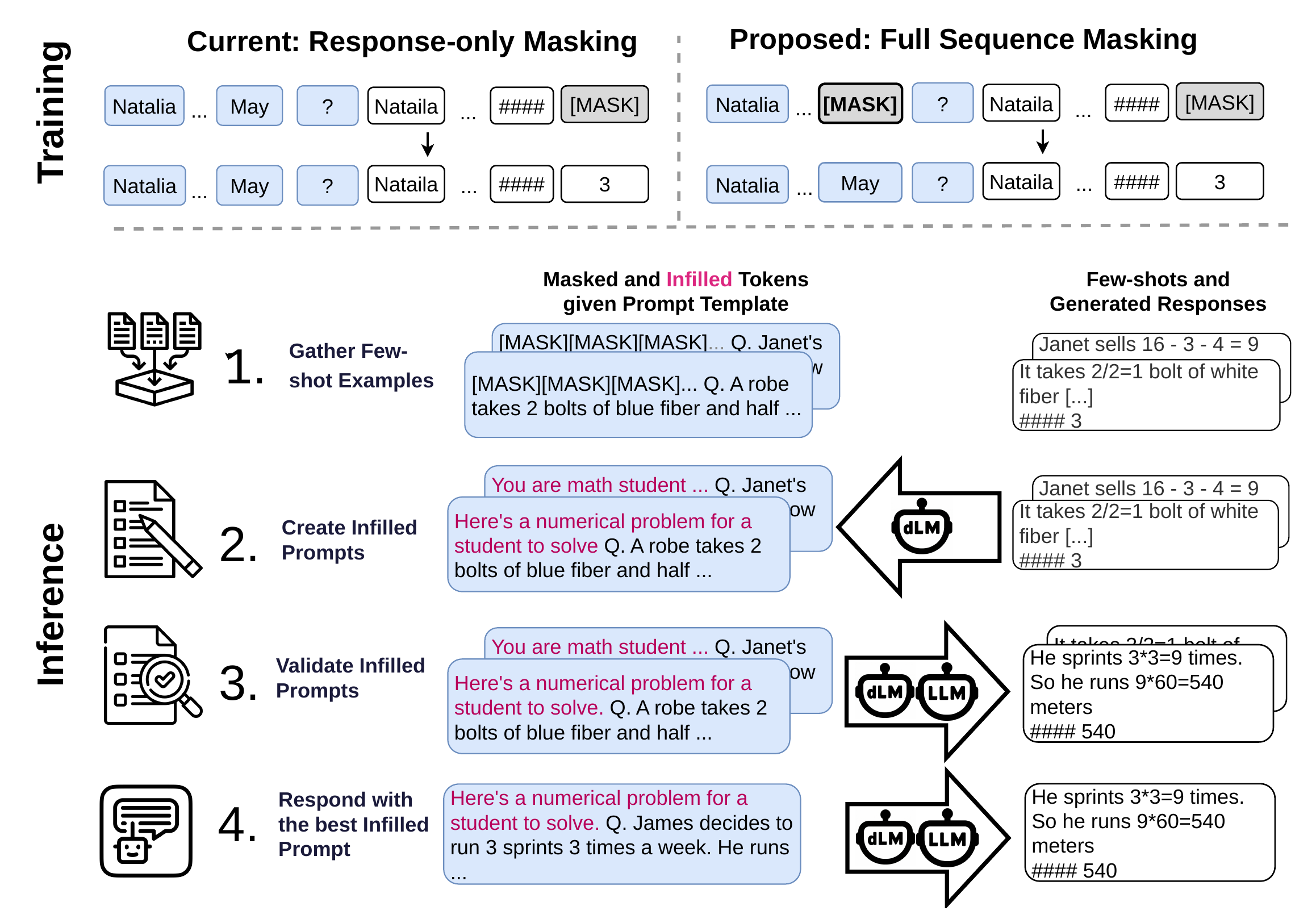}
\caption{Overview of the prompt infilling procedure and the change in training. (1) Gather few-shot examples with prompt templates and reference responses. (2) A diffusion LM (dLM) infills masked tokens in the prompt template, conditioned on the reference responses. (3) Validate infilled prompts by generating responses across all few-shot examples using either a dLM or an LLM. (4) The best infilled prompt is used for final inference on all inputs.}
\label{fig:overview}
\end{figure*}

We address this gap with a simple modification: \emph{full-sequence masking during SFT}, where both prompts and responses are masked. This single change unlocks a model capability that we leverage for a practical application: \emph{prompt infilling}. Given a prompt template with masked regions and a few labeled examples as context, the dLM denoises the masked tokens to produce a complete, task-adapted prompt (Figure~\ref{fig:overview}). The resulting prompts match or surpass manually designed templates, transfer across models, and complement existing prompt optimization methods.

Our main contributions are:

\begin{itemize}
\item We identify that response-only masking during SFT creates a training-inference gap that prevents prompt infilling, and introduce full-sequence masking during SFT as the key ingredient to address this limitation.
\item We introduce a prompt infilling procedure protocol during inference time where infilled prompts match or surpass manually designed templates on multiple datasets, and that these prompts transfer effectively across models.
\item We show that infilled prompts are complementary to existing prompt optimization methods, yielding further gains when combined.
\end{itemize}

\section{Background and Motivation}

We first review how masked diffusion language models operate (\S\ref{sec:mdlm}), then analyze the training-inference gap occurring during SFT stage that prevents prompt infilling (\S\ref{sec:training_inference_gap}), and discuss related work (\S\ref{sec:related_work}).

\subsection{Masked Diffusion Language Models}
\label{sec:mdlm}

Diffusion models generate text by learning to reverse a noising process~\citep{ho2020denoising,song2021scorebased}. Consider a sequence $X = (P, R)$ where $P$ represents the prompt and $R$ the response tokens in $X$. We denote $X_t$ as the sequence at diffusion timestep $t$, where $t=0$ represents clean text and increasing $t$ corresponds to higher noise i.e., more number of masked tokens. Starting from clean text $X_0 = (P_0, R_0)$, the forward diffusion process adds noise to create masked versions $X_t$ at $t$~\citep{austin2021structured,sahoo2024mdlm}.

The model estimates $\mathbb{P}(X_0 | X_t)$ i.e., the probability of recovering clean text from any masked version. During pre-training, models can learn to denoise arbitrary portions of a sequence:
\begin{align}
\mathbb{P}(X_0 | X_t) = \mathbb{P}(P_0, R_0 | P_t, R_t)
\end{align}
where both prompts and responses may be masked. This bidirectional capability distinguishes diffusion models from autoregressive approaches and aligns them with masked language models like BERT~\citep{devlin2018bert} and T5~\citep{raffel2020exploring}, which also perform bidirectional reconstruction. However, diffusion models iterate through multiple denoising steps rather than predicting masked tokens in a single step. This requires the model to encounter diverse masking patterns during training, including when only prompts are masked i.e., $X_{t_p} = (P_t, R_0)$, only responses are masked i.e., $X_{t_r} = (P_0, R_t)$, or both are partially masked.

\subsection{The Training-Inference Gap for Prompt Infilling}
\label{sec:training_inference_gap}

Diffusion models possess an infilling capability: they can denoise and reconstruct any masked portion of a sequence given surrounding context~\citep{sahoo2024mdlm,donahue2020enabling}. This bidirectionality naturally extends to prompt infilling, i.e., when prompt tokens are masked, the model can infill prompts conditioned on desired responses. This capability resembles span corruption in T5~\citep{raffel2020exploring} and infilling in encoder-decoder models~\citep{wang2022self}, but leverages diffusion's iterative refinement process.

However, current SFT practices in LLaDA~\citep{nie2025largelanguagediffusionmodels} and Dream~\citep{ye2025dream7bdiffusionlarge} prevent this capability. During standard SFT, only response tokens are subject to masking while prompts remain completely clean. The model learns to denoise responses given clean prompts, but never encounters scenarios where prompt tokens require denoising. This is a training limitation, not an architectural one: the bidirectional architecture supports prompt infilling, but response-only SFT never provides the necessary training signal (see Appendix~\ref{sec:appendix_training_gap} for formal details).

\subsection{Related Work}
\label{sec:related_work}

\paragraph{Infilling With Diffusion Language Models}

DDOT~\citep{zhang2025flexiblelength} addresses flexible-length text infilling in discrete diffusion models through optimal transport coupling. While DDOT focuses on architectural improvements, our work addresses a training limitation: response-only masking during SFT prevents prompt infilling regardless of architecture. 

\paragraph{Automated Prompt Optimization}

Prompt engineering is crucial for eliciting desired LLM behaviors~\citep{liu2023pre}, but manual engineering is labor-intensive and requires expertise. This has motivated automated approaches: APE~\citep{zhou2022large} generates prompts through iterative refinement, while OPRO~\citep{yang2024large} treats prompt optimization as a meta-learning problem. More recently, DSPy~\citep{khattab2024dspy} provides a framework for compiling declarative language model calls into self-improving pipelines, with methods like GEPA~\citep{agrawal2025gepa} using evolutionary search with frontier models (e.g., GPT-4.1) for prompt optimization. Chain-of-thought prompting~\citep{kojima2022large,wei2022chain} show that well-crafted prompts significantly improve reasoning performance.

However, existing prompt optimization approaches are designed to generate prompts externally with autoregressive models. In contrast, our work enables diffusion models to infer prompts internally from few-shot examples, leveraging inherent infilling capability rather than using larger external models for prompt optimization. Moreover, we show that our infilling approach is complementary to prompt optimization methods (\S\ref{sec:prompt_transfer}), where infilling can further enhance prompts already optimized.

\section{Train with Full Sequence Masking, Infer with Prompt Infilling}
\label{sec:method}

To address the training-inference gap identified above, we introduce full-sequence masking during SFT to unlock prompt infilling (\S\ref{sec:full_seq_masking}). We also describe an optional response-only masking refinement step (\S\ref{sec:response_only_refinement}) and the inference procedure for prompt infilling (\S\ref{sec:infilling_procedure}).

\subsection{Full-Sequence Masking SFT}
\label{sec:full_seq_masking}

In contrast to applying masking only to responses during SFT, we introduce full-sequence masking that masks the entire sequence $X = (P, R)$. For each training example, we sample a masking ratio $t \sim \text{U}(0, 1)$ and randomly select $\lfloor t \times |X| \rfloor$ tokens from the sequence to mask, exposing the model to diverse corruption patterns including prompt-only, response-only, and full-sequence masking. The training loss follows the pretraining phase of LLaDA~\citep{nie2025largelanguagediffusionmodels}, applying cross-entropy loss over all masked tokens:
\begin{align}
\label{eq:full_seq}
\mathcal{L}_{\text{full-seq}} = -\mathbb{E} \left[ \frac{1}{|M_t|} \sum_{i \in M_t} \log \mathbb{P}(X^{i}_{0} | X^{i}_{t}) \right]
\end{align}
where $M_t = \{i : X^{i}_{t} = \texttt{[Mask]}\}$ is the set of all masked positions. This objective enables the model to unmask arbitrary tokens in both prompts and responses.

\paragraph{Optional: Response-Only Masking Refinement}
\label{sec:response_only_refinement}

After full-sequence masking, an optional second fine-tuning stage using response-only masking can improve results on some datasets. This specializes the model for standard generation while preserving prompt infilling capabilities from full-sequence masking. Following \citet{lou2024sedd} and~\citet{nie2025largelanguagediffusionmodels}, this stage uses cross-entropy loss over masked response tokens (Eq.~\ref{eq:sft} in Appendix~\ref{sec:appendix_training_gap}). We show in our experiments that full-sequence masking is crucial on diverse benchmarks, while this refinement step provides additional gains on some tasks.

\subsection{Prompt Infilling during Inference}
\label{sec:infilling_procedure}

At inference time, we leverage the bidirectional denoising capability learned during full-sequence masking (Figure~\ref{fig:overview}). Given few-shot examples with a response $R_0$ and a partially masked prompt template $P_t$, the model infills the masked portions to produce a complete prompt. After multiple prompt candidates are produced, those are validated across all few-shot examples to select the best one. This infilled prompt is then fixed and reused for all test examples. Figure ~\ref{fig:overview} provides the procedural details.
This enables infilling prompts from desired responses without manual prompt engineering. For example, when evaluating with LLM-as-a-Judge on a new dataset with different score scales, the model can infer appropriate scoring rubrics from few-shot examples. 
The infilled prompt is produced once and then applied to all inputs, making the cost of infilling amortized.



\subsection{Empirical Evidence for the Training-Inference Gap}

To confirm that the training-inference gap identified in \S\ref{sec:training_inference_gap} prevents prompt infilling in practice, we evaluate publicly available checkpoints on GSM8K~\citep{cobbe2021training} with three settings: (1) zero-shot prompting, (2) 16-shot in-context learning (ICL), and (3) infilling the first 128 tokens using few-shot examples.
Table~\ref{tab:gsm8k} shows that prompt infilling with public checkpoints consistently hurts performance: LLaDA drops from 73.1\% (ICL) to 64.8\%, and Dream drops from 79.3\% to 54.3\%. Despite strong base performance, these models lack the capability for effective prompt infilling without full-sequence masking during SFT. For example, the public checkpoint produces the following infilled prompt, where most of the masked region is filled with end-of-text tokens rather than meaningful content.

\begin{table}[t]
\centering
\small
\begin{tabular}{llrrr}
\toprule
\textbf{Model} & \textbf{Prompt} & \textbf{\#Shots} & \textbf{EM} \\
\midrule
\multirow{3}{*}{LLaDA}
  & ``Q: ... A: ...'' & 0   & 70.8 \\
  & +ICL               & 16  & \bf 73.1 \\
  & +Prepend Infilled  Tokens & 16  & 69.8 \\
\midrule
\multirow{3}{*}{Dream}
  & ``Q: ... A: ...''  & 0  & \textbf{81.0} & \\
  & +ICL               & 16 & 79.3 & \\
  & +Prepend Infilled Tokens  & 16  & 54.3 & \\
\bottomrule
\end{tabular}

\caption{GSM8K results on public checkpoints (no fine-tuning). In both models, infilled prompts with the publicly available checkpoints decreases the exact match (EM) score (\%).}

\label{tab:gsm8k}
\end{table}

\begin{figure}[t]
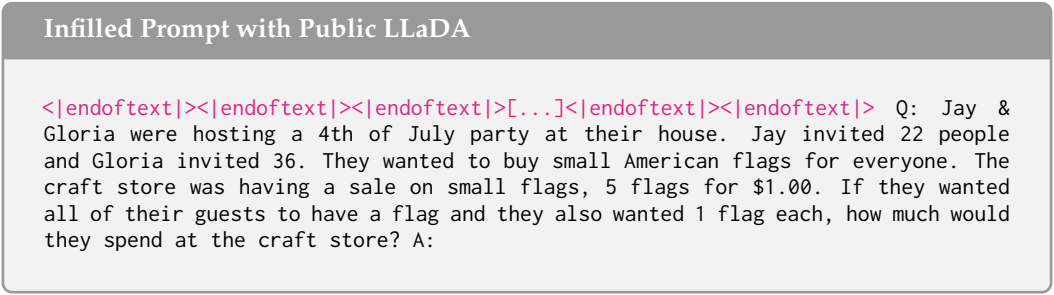

\begin{promptbox}[Infilled Prompt with Public LLaDA]
\textcolor[HTML]{dc267f}{<|endoftext|><|endoftext|><|endoftext|>[...]<|endoftext|><|endoftext|>}
Q: Jay \& Gloria were hosting a 4th of July party at their house. Jay invited 22 people and Gloria invited 36. They wanted to buy small American flags for everyone. The craft store was having a sale on small flags, 5 flags for \$1.00. If they wanted all of their guests to have a flag and they also wanted 1 flag each, how much would they spend at the craft store? A:
\end{promptbox}
\caption{Example prompt with \textcolor[HTML]{dc267f}{infilled tokens} using the public LLaDA checkpoint (\texttt{LLaDA-8B-Instruct}) on GSM8K. All masked prefix tokens are filled with EOS tokens, confirming the training-inference gap for prompt infilling.}
\label{fig:public_infill_example}
\end{figure}

\section{Evaluating Full-Sequence Masking SFT}
\label{sec:main_experiments}

We now evaluate whether full-sequence masking during SFT unlocks prompt infilling capability on LLM-as-a-Judge.
LLM-as-a-Judge presents a complementary challenge: long, multi-section prompts with task descriptions, scoring rubrics, reference answers, and format specifications. 

\subsection{Setup}

\paragraph{Models} We evaluate two masked diffusion language models: \texttt{LLaDA-8B-Instruct}~\citep{nie2025largelanguagediffusionmodels} and \texttt{Dream-v0-Instruct-7B}~\citep{ye2025dream7bdiffusionlarge}.

\paragraph{Training Configurations} We compare three training approaches: (1) \textbf{Response-Only Masking (RO)}: following the standard practice in current diffusion language models; (2) \textbf{Full-Sequence Masking (FS)}: both prompts and completions are subject to masking, enabling the model to learn prompt denoising; (3) \textbf{Full-Sequence + Response-Only (FS+RO)}: full-sequence masking followed by response-only masking refinement. We also evaluate public checkpoints (\textbf{None}) without additional fine-tuning as a baseline.

\paragraph{Data and Prompt Templates} We fine-tune on Feedback Collection~\citep{kim2023prometheus} and evaluate on two human annotated datasets: SummEval~\citep{10.1162/tacl_a_00373} and BigGen-Bench~\citep{kim-etal-2025-biggen}. 
We compare three prompt templates: Judge (from Feedback Collection~\citep{kim2023prometheus}), G-Eval~\citep{liu-etal-2023-g}, and C-gen (Claude-generated), a baseline prompt generated by Claude. We further experiment with more templates in~\S\ref{sec:prompt_transfer_prometheus}.

\subsection{LLM-as-a-Judge Results}
\label{sec:llm_judge_results}

Table~\ref{tab:llm_judge} shows the LLM judge results on SummEval dataset. A key observation is that the ability to infer and generate float numbers (e.g., score ranges like 1.0-5.0, Figure~\ref{fig:public_infill_example}) is a critical contributor to correlation with human judgment.
Comparing different training strategies across LLaDA models reveals a clear progression: the public checkpoint achieves near-zero correlation with Judge Infill ($.087$ Spearman), RO improves to $.477$, and full-sequence masking (FS) reaches $.490$, confirming that it is essential for developing prompt infilling capabilities. Adding response-only refinement (FS+RO) achieves $.535$ on this dataset, showing that the optional second stage can provide further gains without sacrificing the learned prompt infilling abilities. Notably, RO training shows inconsistent infilling performance across prompt types, with high variance in some settings (e.g., C-gen Infill: $.058_{\pm .048}$), suggesting that response-only masking alone is insufficient training for robust prompt infilling.

\begin{table*}[t]
\small
\centering
\setlength{\tabcolsep}{3pt}
\begin{minipage}[t]{0.48\textwidth}
\centering
\begin{tabular}{llccc}
\toprule
\multicolumn{5}{c}{\textbf{LLaDA}} \\
\textbf{Train} & \textbf{Prompt} & \textbf{Pear.} & \textbf{Spear.} & \textbf{Kend.} \\
\midrule
\multirow{5}{*}{None} &
           G-eval                & $.397_{.000}$ & $.289_{.000}$ & $.217_{.000}$ \\
      & C-gen                   & $\underline{.465}_{.000}$ & $\underline{.439}_{.000}$ & $\underline{.339}_{.000}$ \\
      & C-gen Infill & -$.005_{.043}$ & -$.013_{.065}$ & -$.009_{.044}$ \\
      & Judge        & $.308_{.000}$ & $.336_{.000}$ & $.261_{.000}$ \\
      & Judge Infill & $.050_{.063}$ & $.087_{.069}$ & $.064_{.053}$ \\
\cline{1-5}
\multirow{5}{*}{RO} &
           G-Eval                & $.418_{.000}$ & $.482_{.000}$ & $.373_{.000}$ \\
      & C-gen                   & $.454_{.002}$ & $.491_{.002}$ & $\underline{.399}_{.002}$ \\
      & C-gen Infill & $.038_{.025}$ & $.058_{.048}$ & $.042_{.034}$ \\
      & Judge        & $.484_{.000}$ & $.495_{.000}$ & $.391_{.000}$ \\
      & Judge Infill & $\underline{.516}_{.024}$ & $\underline{.504}_{.015}$ & $.396_{.009}$ \\
\cline{1-5}
\multirow{5}{*}{FS} & G-Eval             & $\underline{.453}_{.004}$ & $.436_{.004}$ & $.346_{.003}$ \\
      & C-gen & $.376_{.000}$ & $.404_{.000}$ & $.322_{.000}$ \\
      & C-gen Infill & $.317_{.009}$ & $.345_{.013}$ & $.275_{.010}$ \\
      & Judge     & $.447_{.000}$ & $\underline{.481}_{.000}$ & $\underline{.385}_{.000}$ \\
      & Judge Infill & $.320_{.257}$ & $.465_{.044}$ & $.369_{.035}$ \\
\cline{1-5}
\multirow{5}{*}{FS+RO} & G-Eval             & $.386_{.000}$ & $.480_{.000}$ & $.380_{.000}$ \\
      & C-gen & $.432_{.000}$ & $.449_{.000}$ & $.359_{.000}$ \\
      & C-gen Infill & $.209_{.103}$ & $.159_{.106}$ & $.128_{.083}$ \\
      & Judge     & $.500_{.000}$ & $.507_{.000}$ & $.404_{.000}$ \\
      & Judge Infill & $\mathbf{.546}_{.013}$ & $\mathbf{.535}_{.007}$ & $\mathbf{.423}_{.006}$ \\
\bottomrule
\end{tabular}
\end{minipage}
\hfill
\begin{minipage}[t]{0.48\textwidth}
\centering
\begin{tabular}{llccc}
\toprule
\multicolumn{5}{c}{\textbf{Dream}} \\
\textbf{Train} & \textbf{Prompt} & \textbf{Pear.} & \textbf{Spear.} & \textbf{Kend.} \\
\midrule
\multirow{5}{*}{None} &
           G-eval                & $.247_{.014}$ & $.231_{.030}$ & $.184_{.025}$ \\
      & C-gen                   & $\underline{.292}_{.020}$ & $\underline{.279}_{.023}$ & $\underline{.225}_{.019}$ \\
      & C-gen Infill & $.149_{.157}$ & $.143_{.172}$ & $.112_{.136}$ \\
      & Judge        & $.192_{.001}$ & $.181_{.013}$ & $.136_{.008}$ \\
      & Judge Infill & $.153_{.069}$ & $.160_{.069}$ & $.122_{.050}$ \\
\cline{1-5}
\multirow{5}{*}{RO} &
           G-Eval                & $\underline{.323}_{.019}$ & $\underline{.323}_{.011}$ & $.248_{.008}$ \\
      & C-gen                   & $.264_{.006}$ & $.261_{.005}$ & $.201_{.004}$ \\
      & C-gen Infill & $.247_{.264}$ & $.200_{.318}$ & $.151_{.249}$ \\
      & Judge        & $.285_{.029}$ & $.322_{.013}$ & $\underline{.255}_{.011}$ \\
      & Judge Infill & $.260_{.050}$ & $.267_{.068}$ & $.205_{.053}$ \\
\cline{1-5}
\multirow{5}{*}{FS} & G-Eval             & $.206_{.271}$ & $.338_{.055}$ & $.264_{.042}$ \\
      & C-gen & $.317_{.079}$ & $\underline{.361}_{.003}$ & $\underline{.277}_{.001}$ \\
      & C-gen Infill & $\underline{.344}_{.100}$ & $.328_{.104}$ & $.258_{.081}$ \\
      & Judge     & $.244_{.022}$ & $.230_{.028}$ & $.180_{.022}$ \\
      & Judge Infill & $.220_{.023}$ & $.216_{.019}$ & $.167_{.011}$ \\
\cline{1-5}
\multirow{5}{*}{FS+RO} & G-Eval             & $.312_{.029}$ & $.314_{.027}$ & $.244_{.022}$ \\
      & C-gen & $.326_{.016}$ & $.346_{.030}$ & $.266_{.024}$ \\
      & C-gen Infill & $\mathbf{.461}_{.019}$ & $\mathbf{.454}_{.011}$ & $\mathbf{.344}_{.008}$ \\
      & Judge     & $.227_{.011}$ & $.222_{.007}$ & $.173_{.007}$ \\
      & Judge Infill & $.368_{.047}$ & $.381_{.028}$ & $.300_{.021}$ \\
\bottomrule
\end{tabular}
\end{minipage}

\caption{Model performance comparison with different prompting strategies on SummEval dataset. \textbf{Bold} indicates best performance within each model; \underline{underline} indicates best prompt per (Model, Train) setup.}

\label{tab:llm_judge}
\end{table*}

\begin{figure*}[t]
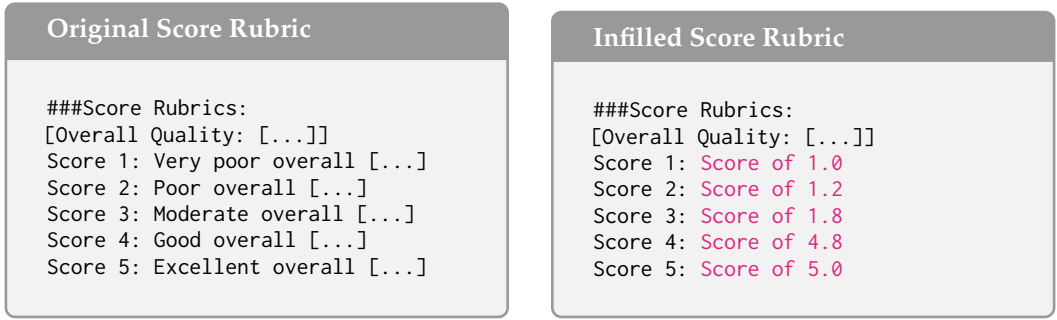

\begin{minipage}[t]{0.48\textwidth}
\begin{promptbox}[Original Score Rubric]
\#\#\#Score Rubrics:\\
{[}Overall Quality: {[}...{]}{]}\\
Score 1: Very poor overall {[}...{]}\\
Score 2: Poor overall {[}...{]}\\
Score 3: Moderate overall {[}...{]}\\
Score 4: Good overall {[}...{]}\\
Score 5: Excellent overall {[}...{]}
\end{promptbox}
\end{minipage}
\hfill
\begin{minipage}[t]{0.48\textwidth}
\begin{promptbox}[Infilled Score Rubric]
\#\#\#Score Rubrics:\\
{[}Overall Quality: {[}...{]}{]}\\
Score 1: \textcolor[HTML]{dc267f}{Score of 1.0}\\
Score 2: \textcolor[HTML]{dc267f}{Score of 1.2}\\
Score 3: \textcolor[HTML]{dc267f}{Score of 1.8}\\
Score 4: \textcolor[HTML]{dc267f}{Score of 4.8}\\
Score 5: \textcolor[HTML]{dc267f}{Score of 5.0}
\end{promptbox}
\end{minipage}
\caption{Example of a score rubric with \textcolor[HTML]{dc267f}{infilled tokens} by the model (right) and original score rubric (left). The model replaces descriptions with non-uniform score values (e.g., 1.2, 1.8, 4.8) inferred from few-shot examples, encouraging float score outputs which results in better human correlation.}
\label{fig:infilled_rubric}
\end{figure*}

The effectiveness of different prompt templates varies significantly across training stages. For LLaDA with FS+RO training, the Judge template consistently outperforms both G-Eval and C-gen prompts, achieving $0.535$ Spearman correlation with infilling compared to $0.480$ (G-Eval) and $0.449$ (C-gen). This superiority likely stems from the Judge template's structured format with explicit task descriptions, scoring rubrics, and reference answers i.e., components that align naturally with the bidirectional denoising patterns learned during full-sequence masking. In contrast, Dream models show more varied behavior: while FS+RO Dream achieves strong performance with C-gen Infill ($.454$ Spearman), its Judge Infill performance ($.381$) lags behind LLaDA, suggesting potential differences in how the two base models internalize structured prompt formats during training (we further analyze this via diffusion perplexity in \S\ref{sec:diffusion_ppl}). The consistent improvement of infilled prompts over oracle prompts in well-trained models (e.g., LLaDA FS+RO: Judge Infill $.535$ vs. Judge oracle $.507$) validates that prompt infilling can discover task-specific optimizations beyond manually designed templates.

\paragraph{BigGen-Bench Evaluation} To further validate generalization beyond SummEval dataset, we evaluate on BigGen-Bench~\citep{kim-etal-2025-biggen}, a more challenging benchmark with 2,780 diverse LLM-as-a-Judge examples. Table~\ref{tab:biggen_and_ppl}a shows results using LLaDA models with the infilled judge template.
The public checkpoint achieves near-zero correlation ($.078$ Spearman, $.068$ Kendall's tau), showing that prompt infilling capability does not emerge from architectural design alone but requires explicit training with full-sequence masking during SFT.
Interestingly, on BigGen-Bench, FS achieves the strongest performance ($.548$ Spearman) with lowest variance ($\pm .026$), surpassing FS+RO ($.525$). This contrasts with SummEval where FS+RO was optimal. The superior FS performance suggests that for this more diverse benchmark, prompt denoising capability is most critical, while RO may introduce overfitting to specific response patterns. The substantially lower variance of FS indicates more robust generalization. RO shows performance ($.406$) significantly better than the public checkpoint but weaker than FS models, confirming that response-only masking provides insufficient signal for robust prompt infilling.

\begin{table*}[t]
\small
\centering
\begin{minipage}[t]{0.52\textwidth}
\centering
\begin{tabular}{lllll}
\toprule
\textbf{Train} & \textbf{Pear.}$\uparrow$ & \textbf{Spear.}$\uparrow$ & \textbf{Kend.}$\uparrow$ & \textbf{PPL}$\downarrow$ \\
\midrule
Public      & -$.007_{.044}$ & $.078_{.058}$ & $.068_{.050}$ & $7.40$ \\
RO     & $.205_{.072}$ & $.406_{.091}$ & $.168_{.079}$ & $7.14$ \\
FS     & $\mathbf{.312}_{.013}$ & $\mathbf{.548}_{.026}$ & $\mathbf{.263}_{.014}$ & $\mathbf{3.73}$ \\
FS+RO   & $.259_{.039}$ & $.525_{.049}$ & $.210_{.043}$ & $3.78$ \\
\bottomrule
\end{tabular}
\vspace{1em}

\centerline{(a) BigGen-Bench (Judge Infill)}
\end{minipage}
\hfill
\begin{minipage}[t]{0.42\textwidth}
\centering
\begin{tabular}{lcc}
\toprule
\textbf{Train} & \textbf{Judge PPL}$\downarrow$ & \textbf{C-gen PPL}$\downarrow$ \\
\midrule
Public    & 11.49 & 9.18 \\
RO   & 13.13 & 9.82 \\
FS   & 5.10 & 12.10 \\
FS+RO & \bf 4.92 & 11.63 \\
\bottomrule
\end{tabular}
\vspace{1em}

\centerline{(b) SummEval PPL by template}
\end{minipage}

\caption{Diffusion perplexity analysis with LLaDA. (a) BigGen-Bench results using Judge Infill prompt across finetuning stages. PPL: diffusion perplexity with gold responses. Full-sequence masking (FS) achieves best performance with lowest variance and lowest perplexity. (b) Diffusion perplexity ($\downarrow$) on SummEval with gold responses (\texttt{[RESULT] \{score\}}) using finetuned and non-finetuned variants of LLaDA-8B-Instruct.}

\label{tab:biggen_and_ppl}
\end{table*}

\subsection{Diffusion Perplexity Analysis on Different Prompt Templates and Finetuning}
\label{sec:diffusion_ppl}
To understand why different prompt templates are optimal on SummEval dataset, we evaluate test set perplexity using the diffusion perplexity (PPL) metric from~\citet{sahoo2024mdlm}, which measures model fit via the negative evidence lower bound (NELBO). See Appendix~\ref{sec:appendix_diffusion_ppl} for the formal definition.

\paragraph{Diffusion Perplexity on SummEval}
Table~\ref{tab:biggen_and_ppl}b shows diffusion perplexity on SummEval evaluation prompts appended with gold-standard responses (\texttt{[RESULT] \{score\}}) using human annotations. FS and FS+RO checkpoints, which were trained on the Judge template format, achieve substantially lower perplexity on Judge-template prompts (5.10 and 4.92) but higher perplexity on C-gen-template prompts (12.10 and 11.63). Conversely, the Public and RO models show lower perplexity on C-gen-template prompts (9.18 and 9.82) but higher on Judge-template prompts (11.49 and 13.13). This confirms the template-dependent evaluation results observed in Table~\ref{tab:llm_judge}.

\paragraph{Diffusion Perplexity on BigGen-Bench} The PPL column in Table~\ref{tab:biggen_and_ppl}a extends this analysis to BigGen-Bench. The same pattern holds: FS and FS+RO achieve substantially lower perplexity (3.73 and 3.78) than Public and RO (7.40 and 7.14), confirming that full-sequence masking is the critical factor. FS achieves slightly lower perplexity than FS+RO, consistent with its stronger evaluation performance. Both FS models achieve lower perplexity on BigGen-Bench than on SummEval (4.92--5.10), likely reflecting the greater diversity of BigGen-Bench prompts.

\section{Diffusion LM as a Prompt Proposal Model for other LMs}
\label{sec:prompt_transfer}

To evaluate whether infilled prompts generalize across 1) models with the same architecture but with different finetuned checkpoints (\S\ref{sec:prompt_transfer_gsm8k}, \S\ref{sec:prompt_transfer_llm_judge}) and 2) to different models (\S\ref{sec:prompt_transfer_prometheus}), we use the infilled prompt from FS+RO during validation and evaluate with other models.

\subsection{Compressing Prompts}
\label{sec:prompt_transfer_gsm8k}

\begin{figure*}[t]
\centering
\begin{subfigure}[t]{0.48\textwidth}
\centering
\includegraphics[width=\textwidth]{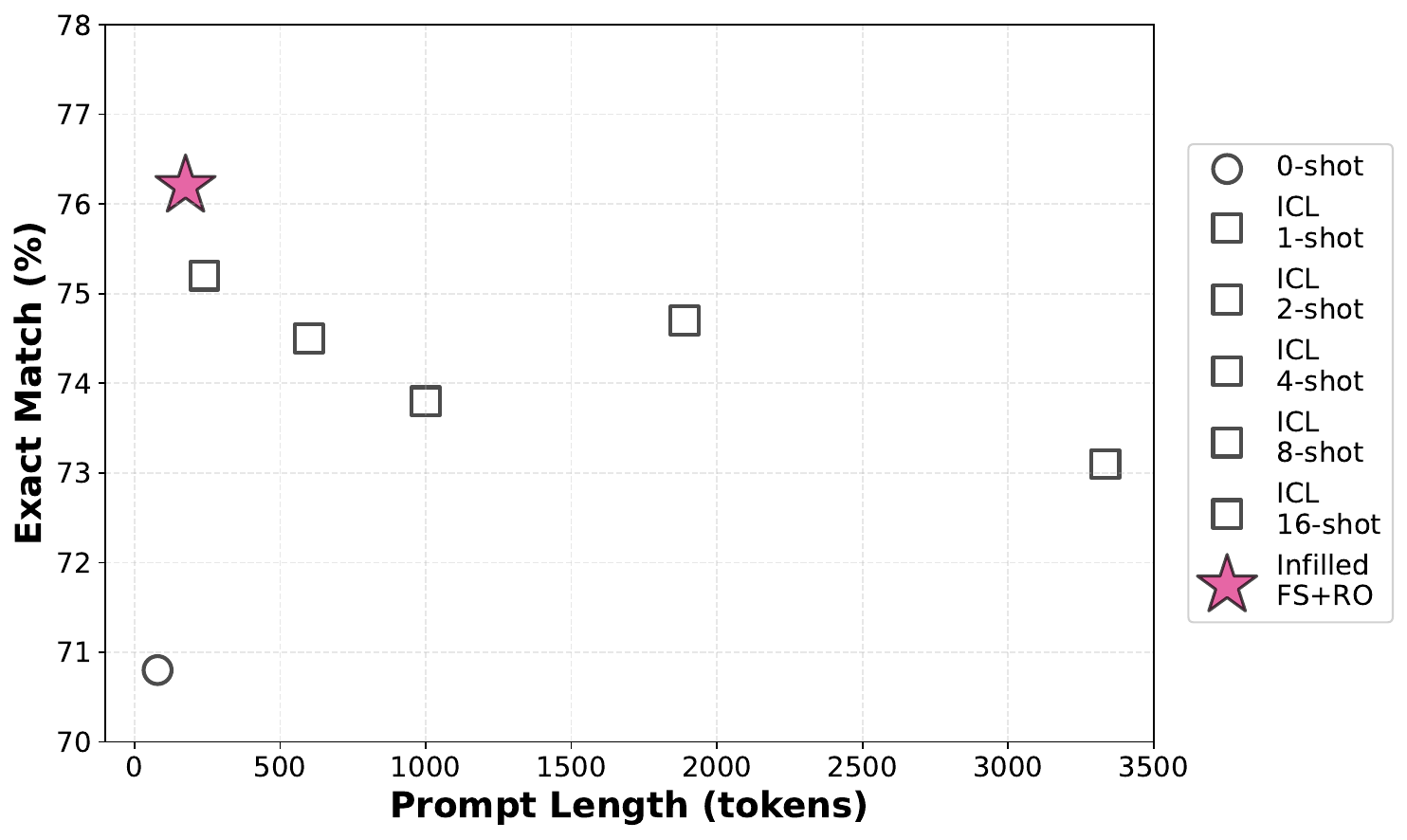}
\caption{Prompt transfer results on GSM8K using LLaDA public checkpoints with the prompt template ``Q:\{question\} A:\{answer\}'' from~\citet{kojima2022large} for in-context learning (ICL).}
\label{tab:gsm8k_prompt_transfer}
\end{subfigure}
\hfill
\begin{subfigure}[t]{0.48\textwidth}
\centering
\includegraphics[width=\textwidth]{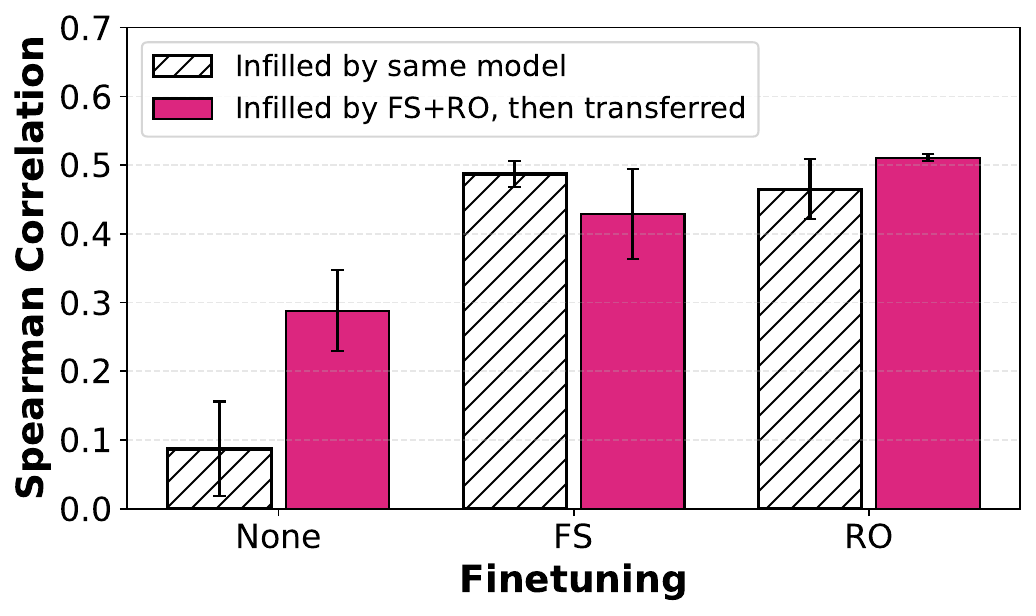}
\caption{Prompt transfer on SummEval using LLaDA models trained with different masking strategies. 
The transferred infilled prompt improves Public and RO models.
}
\label{tab:extracted_prompts}
\end{subfigure}
\caption{Prompt transfer experiments. (a) GSM8K: infilled prompts achieve higher accuracy with fewer tokens than ICL. (b) SummEval: infilled prompts transfer across LLaDA variants with different training configurations.}
\label{fig:prompt_transfer}
\end{figure*}

Figure~\ref{tab:gsm8k_prompt_transfer} shows the effectiveness of prompt transfer on GSM8K. The FS+RO trained model achieves 76.4\% exact match accuracy using inferred prompts, outperforming standard 16-shot in-context learning (73.1\%) while requiring fewer tokens (315.8 vs. 3333.8). This reduction in prompt length with improved performance validates that full-sequence masking enables models to distill essential reasoning patterns from few-shot examples into compact, transferable prompts. We observe similar trends on Dream as well (Appendix~\ref{sec:appendix_gsm8k}). Notably, the public checkpoint's infilling attempt achieves only 64.8\%, confirming that effective prompt infilling requires exposure to full-sequence masking during SFT.

\subsection{Prompt Transfer on LLM-as-a-Judge}
\label{sec:prompt_transfer_llm_judge}
Figure~\ref{tab:extracted_prompts} presents cross-model prompt transfer results using the FS+RO infilled prompt. When transferred to RO, it achieves $.511$ Spearman, surpassing RO's own self-infilling baseline ($.465$), showing that a stronger infilling model can produce prompts that benefit weaker models. The FS model achieves $.429$ with the transferred prompt, slightly below its self-infilling performance ($.487$), suggesting some loss when transferring across training configurations. The public checkpoint improves from $.087$ (self-infilling) to $.288$ with the transferred prompt, a substantial gain but still well below trained models, indicating that while optimized prompts help, the receiving model's training still matters.
These findings suggest that prompts inferred from models trained with full-sequence masking can effectively transfer to other models.

\subsection{Compare and Combine with Prompt Optimization for Autoregressive Models}
\label{sec:prompt_transfer_prometheus}

Automatic prompt optimization methods improve prompts through iterative search over candidate rewrites. However, these approaches rewrite whole prompt: each iteration proposes an entirely new prompt, lacking a mechanism for localized, surgical edits to specific sections (e.g., modifying only the scoring rubric while preserving the task description). In contrast, diffusion-based infilling can mask and refine a targeted region of the prompt while keeping the surrounding context intact, making the two approaches complementary.

Since the structure of prompts produced by these optimization methods is unknown, we use sliding-window (SW) infilling, which masks and infills tokens in the prompt in successive windows and selects the best validated prompt (see Appendix~\ref{sec:appendix_sw_infilling} for configuration details). We evaluate on Prometheus-8x7B v2.0~\citep{kim-etal-2024-prometheus} and Prometheus-13B v1.0~\citep{kim2023prometheus} as the autoregressive judge models. We compare and combine with the following prompt optimization approaches using DSPy~\citep{khattab2024dspy}:

\paragraph{GEPA}~\citep{agrawal2025gepa} is a evolutionary prompt optimization method. It maintains a population of candidate prompts and iteratively generates new variants through mutation and crossover operations, using a frontier model (e.g., GPT-4.1) as the proposal model. Candidates are evaluated on a training set and selected based on task performance. GEPA rewrites prompts holistically at each generation step which can hurt the task accuracy.

\paragraph{COPRO}~\citep{opsahl-ong-etal-2024-optimizing,khattab2024dspy} extends OPRO~\citep{yang2024large} to multi-stage settings. It 1) generates multiple prompt variations using an LLM, 2) evaluates each candidate on training data, 3) uses the best candidates to generate improved versions, and 4) repeats for a fixed number of iterations. Unlike GEPA's population-based evolutionary search, COPRO follows a single hill-climbing trajectory. Like GEPA, COPRO rewrites prompts in their entirety at each step.

\begin{table}[t]
\small
\centering
\begin{tabular}{lllccc}
\toprule
\textbf{Eval Model} & \textbf{Method} & \textbf{Proposal} & \textbf{Pear.} & \textbf{Spear.} & \textbf{Kend.} \\
\midrule
\multirow{5}{*}{Prometheus 8x7B}
  & Baseline & -- & $.428_{.000}$ & $.410_{.000}$ & $.372_{.000}$ \\
  & COPRO & 8x7B & $.465_{.098}$ & $.447_{.073}$ & $.355_{.061}$ \\
  & \textbf{+ Infill} & \textbf{LLaDA 8B} & $\mathbf{.539}_{.020}$ & $\mathbf{.490}_{.010}$ & $\mathbf{.394}_{.011}$ \\
  & GEPA & 8x7B & $.377_{.169}$ & $.356_{.147}$ & $.330_{.136}$ \\
  & + Infill & LLaDA 8B & $.401_{.148}$ & $.391_{.133}$ & $.315_{.108}$ \\
\midrule
\multirow{5}{*}{Prometheus 13B}
  & Baseline & -- & $.271_{.000}$ & $.248_{.000}$ & $.216_{.000}$ \\
  & COPRO & Llama 8B & $.312_{.040}$ & $.306_{.028}$ & $.235_{.023}$ \\
  & \textbf{+ Infill} & \textbf{LLaDA 8B} & $\mathbf{.314}_{.039}$ & $\mathbf{.318}_{.022}$ & $\mathbf{.243}_{.017}$ \\
  & GEPA & Llama 8B & $.131_{.030}$ & $.117_{.046}$ & $.104_{.041}$ \\
  & GEPA & Qwen 14B & $.132_{.069}$ & $.126_{.069}$ & $.114_{.061}$ \\
\bottomrule
\end{tabular}

\caption{Effect of LLaDA infilling on COPRO and GEPA prompts across Prometheus models (8x7B v2.0 and 13B v1.0) on SummEval (16-shot). Infilling improves both methods, but is far more effective on COPRO: variance largely drops ($.073 \rightarrow .010$) while GEPA barely changes ($.147 \rightarrow .133$). COPRO + Infill on 8x7B achieves the best result overall.}

\label{tab:cross_eval_model}
\end{table}

\paragraph{Infilling Effectiveness Combined with Prompt Optimization}
Table~\ref{tab:cross_eval_model} compares the effect of LLaDA infilling on COPRO- and GEPA-optimized prompts across Prometheus evaluation models. Infilling consistently improves both methods on 8x7B, but with different variance: COPRO + Infill reduces standard deviation ($.073 \rightarrow .010$) while GEPA + Infill barely changes it ($.147 \rightarrow .133$). This gap arises because of relatively smaller number of shots (i.e., 16-shots) compared to the experiment setting in~\citet{agrawal2025gepa}, and as a result, COPRO's hill-climbing produces prompts amenable to targeted refinement, whereas GEPA can produce overfitted prompts that infilling cannot fully rescue due to smaller number of shots (e.g., one GEPA offset collapses from $.455$ to $.188$). On 13B, GEPA degrades below baseline regardless of reflection model, while COPRO + Infill still improves by $+.070$. Across both eval models, infilling can improve the quality of the base prompt.

\section{Conclusion}
\label{sec:conclusion}

We identified a limitation in current widely-used masked diffusion language models: response-only masking during supervised fine-tuning constrains their infilling capabilities for prompts, despite their bidirectional architecture theoretically supporting such operations. We show that this is due to training practices that never expose models to prompt infilling.
To address this training limitation, we applied full-sequence masking during SFT, where both prompts and responses are masked. This simple change to the training paradigm unlocks prompt infilling capabilities that were architecturally possible but never learned.
Our results show that the training paradigm, not architectural limitations, is the primary bottleneck for prompt infilling in diffusion models. This finding has broader implications: many capabilities we assume require architectural innovations may simply require exposure to appropriate training signals.

%

%
%

\bibliography{colm2026_conference}

\begin{thebibliography}{27}
\providecommand{\natexlab}[1]{#1}
\providecommand{\url}[1]{\texttt{#1}}
\expandafter\ifx\csname urlstyle\endcsname\relax
  \providecommand{\doi}[1]{doi: #1}\else
  \providecommand{\doi}{doi: \begingroup \urlstyle{rm}\Url}\fi

\bibitem[Agrawal et~al.(2025)Agrawal, Tan, Soylu, Ziems, Khare, Opsahl-Ong, Singhvi, Shandilya, Ryan, Jiang, Potts, Sen, Dimakis, Stoica, Klein, Zaharia, and Khattab]{agrawal2025gepa}
Lakshya~A Agrawal, Shangyin Tan, Dilara Soylu, Noah Ziems, Rishi Khare, Krista Opsahl-Ong, Arnav Singhvi, Herumb Shandilya, Michael~J Ryan, Meng Jiang, Christopher Potts, Koushik Sen, Alexandros~G. Dimakis, Ion Stoica, Dan Klein, Matei Zaharia, and Omar Khattab.
\newblock Gepa: Reflective prompt evolution can outperform reinforcement learning, 2025.
\newblock URL \url{https://arxiv.org/abs/2507.19457}.

\bibitem[Austin et~al.(2021)Austin, Johnson, Ho, Tarlow, and van~den Berg]{austin2021structured}
Jacob Austin, Daniel~D Johnson, Jonathan Ho, Daniel Tarlow, and Rianne van~den Berg.
\newblock Structured denoising diffusion models in discrete state-spaces.
\newblock In \emph{Advances in Neural Information Processing Systems}, volume~34, pp.\  17981--17993, 2021.
\newblock URL \url{https://arxiv.org/abs/2107.03006}.

\bibitem[Cobbe et~al.(2021)Cobbe, Kosaraju, Bavarian, Chen, Jun, Kaiser, Plappert, Tworek, Hilton, Nakano, Hesse, and Schulman]{cobbe2021training}
Karl Cobbe, Vineet Kosaraju, Mohammad Bavarian, Mark Chen, Heewoo Jun, Lukasz Kaiser, Matthias Plappert, Jerry Tworek, Jacob Hilton, Reiichiro Nakano, Christopher Hesse, and John Schulman.
\newblock Training verifiers to solve math word problems, 2021.
\newblock URL \url{https://arxiv.org/abs/2110.14168}.

\bibitem[Devlin et~al.(2019)Devlin, Chang, Lee, and Toutanova]{devlin2018bert}
Jacob Devlin, Ming-Wei Chang, Kenton Lee, and Kristina Toutanova.
\newblock {BERT}: Pre-training of deep bidirectional transformers for language understanding.
\newblock In \emph{Proceedings of the 2019 Conference of the North American Chapter of the Association for Computational Linguistics: Human Language Technologies, Volume 1 (Long and Short Papers)}, pp.\  4171--4186, 2019.
\newblock URL \url{https://arxiv.org/abs/1810.04805}.

\bibitem[Donahue et~al.(2020)Donahue, Lee, and Liang]{donahue2020enabling}
Chris Donahue, Mina Lee, and Percy Liang.
\newblock Enabling language models to fill in the blanks.
\newblock In \emph{Proceedings of the 58th Annual Meeting of the Association for Computational Linguistics}, pp.\  2492--2501, 2020.
\newblock URL \url{https://arxiv.org/abs/2005.05339}.

\bibitem[Fabbri et~al.(2021)Fabbri, Kryściński, McCann, Xiong, Socher, and Radev]{10.1162/tacl_a_00373}
Alexander~R. Fabbri, Wojciech Kryściński, Bryan McCann, Caiming Xiong, Richard Socher, and Dragomir Radev.
\newblock Summeval: Re-evaluating summarization evaluation.
\newblock \emph{Transactions of the Association for Computational Linguistics}, 9:\penalty0 391--409, 04 2021.
\newblock ISSN 2307-387X.
\newblock \doi{10.1162/tacl_a_00373}.
\newblock URL \url{https://doi.org/10.1162/tacl_a_00373}.

\bibitem[Ho et~al.(2020)Ho, Jain, and Abbeel]{ho2020denoising}
Jonathan Ho, Ajay Jain, and Pieter Abbeel.
\newblock Denoising diffusion probabilistic models.
\newblock In \emph{Advances in Neural Information Processing Systems}, volume~33, pp.\  6840--6851, 2020.
\newblock URL \url{https://arxiv.org/abs/2006.11239}.

\bibitem[Jiang et~al.(2020)Jiang, Bordia, Zhong, Dognin, Singh, and Bansal]{jiang-etal-2020-hover}
Yichen Jiang, Shikha Bordia, Zheng Zhong, Charles Dognin, Maneesh Singh, and Mohit Bansal.
\newblock {H}o{V}er: A dataset for many-hop fact extraction and claim verification.
\newblock In Trevor Cohn, Yulan He, and Yang Liu (eds.), \emph{Findings of the Association for Computational Linguistics: EMNLP 2020}, pp.\  3441--3460, Online, November 2020. Association for Computational Linguistics.
\newblock \doi{10.18653/v1/2020.findings-emnlp.309}.
\newblock URL \url{https://aclanthology.org/2020.findings-emnlp.309/}.

\bibitem[Khattab et~al.(2024)Khattab, Singhvi, Maheshwari, Zhang, Santhanam, Vardhamanan, Haq, Sharma, Joshi, Moazam, Miller, Zaharia, and Potts]{khattab2024dspy}
Omar Khattab, Arnav Singhvi, Paridhi Maheshwari, Zhiyuan Zhang, Keshav Santhanam, Sri Vardhamanan, Saiful Haq, Ashutosh Sharma, Thomas~T. Joshi, Hanna Moazam, Heather Miller, Matei Zaharia, and Christopher Potts.
\newblock Dspy: Compiling declarative language model calls into self-improving pipelines.
\newblock In \emph{The Twelfth International Conference on Learning Representations}, 2024.
\newblock URL \url{https://openreview.net/forum?id=sY4VZ0pFSo}.

\bibitem[Kim et~al.(2023)Kim, Shin, Cho, Jang, Longpre, Lee, Yun, Shin, Kim, Thorne, and Seo]{kim2023prometheus}
Seungone Kim, Jamin Shin, Yejin Cho, Joel Jang, Shayne Longpre, Hwaran Lee, Sangdoo Yun, Seongjin Shin, Sungdong Kim, James Thorne, and Minjoon Seo.
\newblock Prometheus: Inducing fine-grained evaluation capability in language models, 2023.

\bibitem[Kim et~al.(2024)Kim, Suk, Longpre, Lin, Shin, Welleck, Neubig, Lee, Lee, and Seo]{kim-etal-2024-prometheus}
Seungone Kim, Juyoung Suk, Shayne Longpre, Bill~Yuchen Lin, Jamin Shin, Sean Welleck, Graham Neubig, Moontae Lee, Kyungjae Lee, and Minjoon Seo.
\newblock Prometheus 2: An open source language model specialized in evaluating other language models.
\newblock In Yaser Al-Onaizan, Mohit Bansal, and Yun-Nung Chen (eds.), \emph{Proceedings of the 2024 Conference on Empirical Methods in Natural Language Processing}, pp.\  4334--4353, Miami, Florida, USA, November 2024. Association for Computational Linguistics.
\newblock \doi{10.18653/v1/2024.emnlp-main.248}.
\newblock URL \url{https://aclanthology.org/2024.emnlp-main.248/}.

\bibitem[Kim et~al.(2025)Kim, Suk, Cho, Longpre, Kim, Yoon, Son, Cho, Shafayat, Baek, Park, Hwang, Jo, Cho, Shin, Lee, Oh, Lee, Ho, Joo, Ko, Lee, Chae, Shin, Jang, Ye, Lin, Welleck, Neubig, Lee, Lee, and Seo]{kim-etal-2025-biggen}
Seungone Kim, Juyoung Suk, Ji~Yong Cho, Shayne Longpre, Chaeeun Kim, Dongkeun Yoon, Guijin Son, Yejin Cho, Sheikh Shafayat, Jinheon Baek, Sue~Hyun Park, Hyeonbin Hwang, Jinkyung Jo, Hyowon Cho, Haebin Shin, Seongyun Lee, Hanseok Oh, Noah Lee, Namgyu Ho, Se~June Joo, Miyoung Ko, Yoonjoo Lee, Hyungjoo Chae, Jamin Shin, Joel Jang, Seonghyeon Ye, Bill~Yuchen Lin, Sean Welleck, Graham Neubig, Moontae Lee, Kyungjae Lee, and Minjoon Seo.
\newblock The {B}i{GG}en bench: A principled benchmark for fine-grained evaluation of language models with language models.
\newblock In Luis Chiruzzo, Alan Ritter, and Lu~Wang (eds.), \emph{Proceedings of the 2025 Conference of the Nations of the Americas Chapter of the Association for Computational Linguistics: Human Language Technologies (Volume 1: Long Papers)}, pp.\  5877--5919, Albuquerque, New Mexico, April 2025. Association for Computational Linguistics.
\newblock ISBN 979-8-89176-189-6.
\newblock \doi{10.18653/v1/2025.naacl-long.303}.
\newblock URL \url{https://aclanthology.org/2025.naacl-long.303/}.

\bibitem[Kojima et~al.(2022)Kojima, Gu, Reid, Matsuo, and Iwasawa]{kojima2022large}
Takeshi Kojima, Shixiang~Shane Gu, Machel Reid, Yutaka Matsuo, and Yusuke Iwasawa.
\newblock Large language models are zero-shot reasoners.
\newblock \emph{Advances in neural information processing systems}, 35:\penalty0 22199--22213, 2022.

\bibitem[Liu et~al.(2023{\natexlab{a}})Liu, Yuan, Fu, Jiang, Hayashi, and Neubig]{liu2023pre}
Pengfei Liu, Weizhe Yuan, Jinlan Fu, Zhengbao Jiang, Hiroaki Hayashi, and Graham Neubig.
\newblock Pre-train, prompt, and predict: A systematic survey of prompting methods in natural language processing.
\newblock \emph{ACM Computing Surveys}, 55\penalty0 (9):\penalty0 1--35, 2023{\natexlab{a}}.
\newblock \doi{10.1145/3560815}.
\newblock URL \url{https://arxiv.org/abs/2107.13586}.

\bibitem[Liu et~al.(2023{\natexlab{b}})Liu, Iter, Xu, Wang, Xu, and Zhu]{liu-etal-2023-g}
Yang Liu, Dan Iter, Yichong Xu, Shuohang Wang, Ruochen Xu, and Chenguang Zhu.
\newblock {G}-eval: {NLG} evaluation using gpt-4 with better human alignment.
\newblock In \emph{Proceedings of the 2023 Conference on Empirical Methods in Natural Language Processing}, pp.\  2511--2522, 2023{\natexlab{b}}.
\newblock \doi{10.18653/v1/2023.emnlp-main.153}.
\newblock URL \url{https://aclanthology.org/2023.emnlp-main.153/}.

\bibitem[Lou et~al.(2024)Lou, Meng, and Ermon]{lou2024sedd}
Aaron Lou, Chenlin Meng, and Stefano Ermon.
\newblock Discrete diffusion modeling by estimating the ratios of the data distribution, 2024.
\newblock URL \url{https://arxiv.org/abs/2310.16834}.
\newblock ICML 2024 Best Paper.

\bibitem[Nie et~al.(2025)Nie, Zhu, You, Zhang, Ou, Hu, Zhou, Lin, Wen, and Li]{nie2025largelanguagediffusionmodels}
Shen Nie, Fengqi Zhu, Zebin You, Xiaolu Zhang, Jingyang Ou, Jun Hu, Jun Zhou, Yankai Lin, Ji-Rong Wen, and Chongxuan Li.
\newblock Large language diffusion models, 2025.
\newblock URL \url{https://arxiv.org/abs/2502.09992}.

\bibitem[Opsahl-Ong et~al.(2024)Opsahl-Ong, Ryan, Purtell, Broman, Potts, Zaharia, and Khattab]{opsahl-ong-etal-2024-optimizing}
Krista Opsahl-Ong, Michael~J Ryan, Josh Purtell, David Broman, Christopher Potts, Matei Zaharia, and Omar Khattab.
\newblock Optimizing instructions and demonstrations for multi-stage language model programs.
\newblock In \emph{Proceedings of the 2024 Conference on Empirical Methods in Natural Language Processing}, pp.\  9340--9366, 2024.
\newblock \doi{10.18653/v1/2024.emnlp-main.525}.
\newblock URL \url{https://aclanthology.org/2024.emnlp-main.525/}.

\bibitem[Raffel et~al.(2020)Raffel, Shazeer, Roberts, Lee, Narang, Matena, Zhou, Li, and Liu]{raffel2020exploring}
Colin Raffel, Noam Shazeer, Adam Roberts, Katherine Lee, Sharan Narang, Michael Matena, Yanqi Zhou, Wei Li, and Peter~J Liu.
\newblock Exploring the limits of transfer learning with a unified text-to-text transformer.
\newblock \emph{Journal of Machine Learning Research}, 21\penalty0 (140):\penalty0 1--67, 2020.
\newblock URL \url{https://arxiv.org/abs/1910.10683}.

\bibitem[Sahoo et~al.(2024)Sahoo, Arriola, Schiff, Gokaslan, Marroquin, Chiu, Rush, and Kuleshov]{sahoo2024mdlm}
Subham~Sekhar Sahoo, Marianne Arriola, Yair Schiff, Aaron Gokaslan, Edgar Marroquin, Justin~T Chiu, Alexander Rush, and Volodymyr Kuleshov.
\newblock Simple and effective masked diffusion language models, 2024.
\newblock URL \url{https://arxiv.org/abs/2406.07524}.

\bibitem[Song et~al.(2021)Song, Sohl-Dickstein, Kingma, Kumar, Ermon, and Poole]{song2021scorebased}
Yang Song, Jascha Sohl-Dickstein, Diederik~P Kingma, Abhishek Kumar, Stefano Ermon, and Ben Poole.
\newblock Score-based generative modeling through stochastic differential equations.
\newblock In \emph{International Conference on Learning Representations}, 2021.
\newblock URL \url{https://arxiv.org/abs/2011.13456}.

\bibitem[Wang et~al.(2022)Wang, Kordi, Mishra, Liu, Smith, Khashabi, and Hajishirzi]{wang2022self}
Yizhong Wang, Yeganeh Kordi, Swaroop Mishra, Alisa Liu, Noah~A Smith, Daniel Khashabi, and Hannaneh Hajishirzi.
\newblock Self-instruct: Aligning language models with self-generated instructions, 2022.
\newblock URL \url{https://arxiv.org/abs/2212.10560}.

\bibitem[Wei et~al.(2022)Wei, Wang, Schuurmans, Bosma, Xia, Chi, Le, Zhou, et~al.]{wei2022chain}
Jason Wei, Xuezhi Wang, Dale Schuurmans, Maarten Bosma, Fei Xia, Ed~Chi, Quoc~V Le, Denny Zhou, et~al.
\newblock Chain-of-thought prompting elicits reasoning in large language models.
\newblock In \emph{Advances in Neural Information Processing Systems}, volume~35, pp.\  24824--24837, 2022.
\newblock URL \url{https://arxiv.org/abs/2201.11903}.

\bibitem[Yang et~al.(2024)Yang, Wang, Lu, Liu, Le, Zhou, and Chen]{yang2024large}
Chengrun Yang, Xuezhi Wang, Yifeng Lu, Hanxiao Liu, Quoc~V Le, Denny Zhou, and Xinyun Chen.
\newblock Large language models as optimizers.
\newblock In \emph{The Twelfth International Conference on Learning Representations}, 2024.
\newblock URL \url{https://openreview.net/forum?id=Bb4VGOWELI}.

\bibitem[Ye et~al.(2025)Ye, Xie, Zheng, Gao, Wu, Jiang, Li, and Kong]{ye2025dream7bdiffusionlarge}
Jiacheng Ye, Zhihui Xie, Lin Zheng, Jiahui Gao, Zirui Wu, Xin Jiang, Zhenguo Li, and Lingpeng Kong.
\newblock Dream 7b: Diffusion large language models, 2025.
\newblock URL \url{https://arxiv.org/abs/2508.15487}.

\bibitem[Zhang et~al.(2025)Zhang, Sivakumar, Tang, and Thomas]{zhang2025flexiblelength}
Andrew Zhang, Anushka Sivakumar, Chia-Wei Tang, and Chris Thomas.
\newblock Flexible-length text infilling for discrete diffusion models.
\newblock In \emph{Proceedings of the 2025 Conference on Empirical Methods in Natural Language Processing}, Suzhou, China, November 2025. Association for Computational Linguistics.
\newblock URL \url{https://aclanthology.org/2025.emnlp-main.1597/}.

\bibitem[Zhou et~al.(2022)Zhou, Muresanu, Han, Paster, Pitis, Chan, and Ba]{zhou2022large}
Yongchao Zhou, Andrei~Ioan Muresanu, Ziwen Han, Keiran Paster, Silviu Pitis, Harris Chan, and Jimmy Ba.
\newblock Large language models are human-level prompt engineers, 2022.
\newblock URL \url{https://arxiv.org/abs/2211.01910}.

\end{thebibliography}
\bibliographystyle{colm2026_conference}

\appendix

\section{Full Prompt Transfer Results}
\label{sec:appendix_prompt_transfer}

\begin{figure}[h]
\centering
\includegraphics[width=0.90\textwidth]{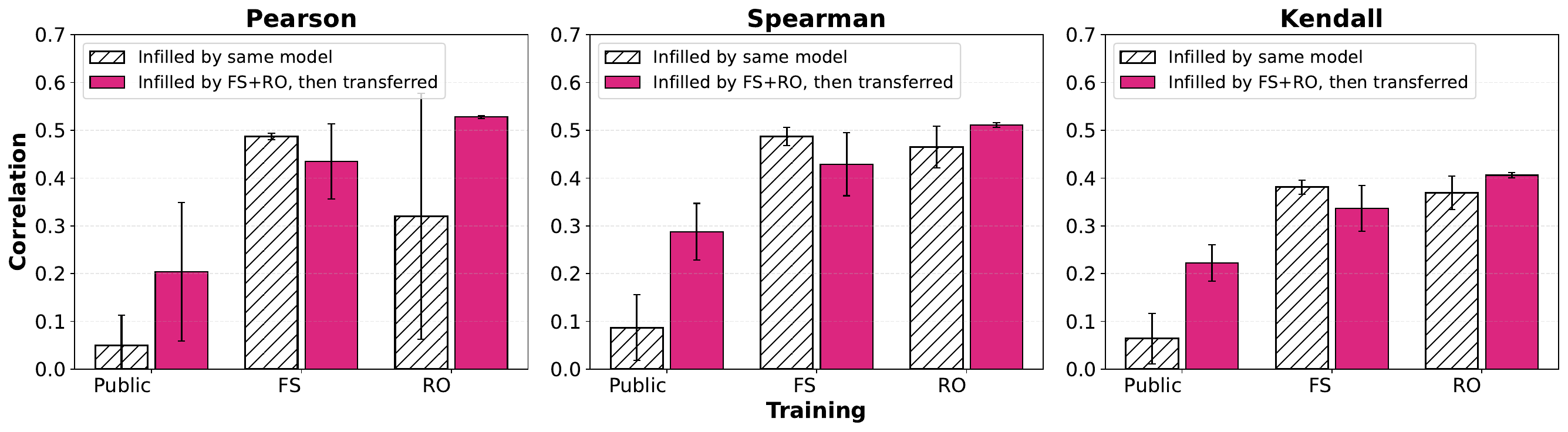}
\caption{Full prompt transfer evaluation on SummEval using LLaDA models trained with different configurations (mean $\pm$ std), showing Pearson, Spearman, and Kendall correlations. Baseline results from Table~\ref{tab:llm_judge} showing Judge Infilled prompts results across different training stages. The transferred prompt improves on Public and RO models showing that the infilled prompts transfer across different LLaDA variants.}
\label{fig:llm_judge_transfer_full}
\end{figure}

\section{Diffusion Perplexity Definition}
\label{sec:appendix_diffusion_ppl}

Following~\citet{sahoo2024mdlm}, we denote $\sigma(t)$ as the noise schedule controlling the mask tokens in a sequence at timestep $t$: higher values of $\sigma$ correspond to more aggressive masking.
We denote $\alpha_t = \exp(-\sigma(t))$ as the probability of a token remaining unmasked at time $t$ (i.e., the survival probability), where $\alpha'_t$ represents the time derivative of $\alpha_t$.
Diffusion perplexity (PPL) is defined using the negative evidence lower bound (NELBO) for a sequence $X^{1:L}$ of length $L$ by
\begin{equation}
\text{PPL}(X) = \exp(\mathcal{L}_{\text{NELBO}}/L)
\end{equation}
NELBO for a sequence $X^{1:L}$ of length $L$ is:
\begin{equation}
\mathcal{L}_{\text{NELBO}} = -\mathbb{E} \int_{t=0}^{t=1} \frac{\alpha'_t}{1-\alpha_t} \sum_{i=1}^{L} \log \mathbb{P}(X^{i}_t | X^{i}_0) \mathrm{d}t
\end{equation}
where $\frac{\alpha'_t}{1-\alpha_t}$ weights each timestep loss, equivalent to weighted version of Eq.~(\ref{eq:full_seq}). In practice, we compute a Monte Carlo estimate $\widehat{\mathcal{L}}_{\text{NELBO}}$ by sampling timesteps and compute diffusion perplexity $\text{PPL}$ as $\exp(\widehat{\mathcal{L}}_{\text{NELBO}}/L)$.
To compute $\widehat{\mathcal{L}}_{\text{NELBO}}$, we sample $t \sim \text{U}(0,1)$ with the noise schedule $\sigma(t) = t\sigma_{\max}$ where $\sigma_{\max}$ is set to $10$.

\section{Training Hyperparameters}
\label{sec:appendix_hyperparams}

Table~\ref{tab:training_hyperparams} summarizes the training hyperparameters for full-sequence masking (FS) and conventional response masking (RO).

\begin{table}[h]
\centering
\begin{tabular}{lp{0.49\columnwidth}}
\toprule
\textbf{Hyperparameter} & \textbf{Value} \\
\midrule
Learning rate (LLaDA) & $1 \times 10^{-5}$ \\
Learning rate (Dream) & $2 \times 10^{-6}$ \\
LR schedule & Cosine decay \\
\midrule
Batch size (LLaDA) & 1 (32 grad.\ accum.\ steps) \\
Batch size (Dream) & 2 (64 grad.\ accum.\ steps) \\
Training epochs (FS) & 8 \\
Training epochs (RO) & 4 \\
\midrule
Warmup & 500 steps (linear) \\
Optimizer & AdamW ($\beta_1\!=\!0.9$, $\beta_2\!=\!0.999$, $\epsilon\!=\!10^{-8}$) \\
Weight decay & $0.01$ \\
Gradient clipping & Max norm $1.0$ \\
\bottomrule
\end{tabular}

\caption{Training hyperparameters for full-sequence masking (FS) and response-only masking (RO).}

\label{tab:training_hyperparams}
\end{table}

\section{Inference Configuration}
\label{sec:appendix_inference}

Table~\ref{tab:inference_config} summarizes the inference configuration.

\begin{table}[h]
\centering
\begin{tabular}{ll}
\toprule
\textbf{Parameter} & \textbf{Value} \\
\midrule
Diffusion steps & $128$ \\
Response length & $128$ \\
Temperature (LLaDA) & $0.8$ \\
Temperature (Dream) & $0.1$ \\
\bottomrule
\end{tabular}

\caption{Inference configuration.}

\label{tab:inference_config}
\end{table}

\section{Sliding-Window Infilling Configuration}
\label{sec:appendix_sw_infilling}

For combining prompt infilling with prompt optimization methods (\S\ref{sec:prompt_transfer_prometheus}), we use sliding-window (SW) infilling since the structure of optimized prompts is unknown. The sliding window moves across the prompt, masking and infilling tokens in successive windows, and selects the best validated prompt. We tuned the window size, stride, and mask size on the validation set. The best performing configuration is window = 8 tokens, stride = 4 tokens, and mask = 8 tokens.

\section{Dataset Statistics}
\label{sec:appendix_dataset_stats}

We provide detailed statistics for all datasets used in our experiments:

\paragraph{Training Data} We fine-tune models on Feedback Collection~\citep{kim2023prometheus}, which contains 95,000 train examples and 5,000 dev examples for LLM-as-a-Judge tasks. Each example includes an instruction, response, reference answer, evaluation criteria, and score descriptions. We use the train split for both FS and RO fine-tuning.

\paragraph{Evaluation Datasets}
\begin{itemize}
    \item \textbf{GSM8K}~\citep{cobbe2021training}: Grade school math reasoning dataset with 7,099 train examples, 374 validation examples, and 1,319 test examples.
    \item \textbf{HoVer}~\citep{jiang-etal-2020-hover}: Multi-hop fact verification dataset. We use a subset for evaluation with 3 hops and 7 documents per search, averaging results over 3 runs with 16 few-shot examples.
    \item \textbf{SummEval}~\citep{10.1162/tacl_a_00373}: Summarization evaluation dataset with human annotations, containing 160 dev examples and 1,440 test examples. We evaluate on the test set containing multiple summarization systems' outputs across different source documents.
    \item \textbf{BigGen-Bench}~\citep{kim-etal-2025-biggen}: Large-scale benchmark for generative tasks with 2,780 diverse LLM-as-a-Judge evaluation examples. We use this for out-of-domain evaluation to test generalization.
\end{itemize}

\section{The Training-Inference Gap: Formal Details}
\label{sec:appendix_training_gap}

During supervised fine-tuning, given a prompt-response pair $X_0 = (P_0, R_0)$, a masking ratio $t$ is sampled uniformly from the interval $(0, 1]$ i.e., $t \sim \text{U}(0, 1]$. Given the sampled ratio $t$, each token $R_0^i$ in the response is independently masked with probability $t$. This produces a masked response at time $t$ i.e., $R_t$ where, in expectation, a fraction $t$ of tokens are replaced with the mask token. 
Within a single training step, all tokens share the same masking probability $t$.

The prompt $P_0$ remains completely clean: $X_{t_r} = (P_0, R_t)$ where $P_0$ is never masked. The SFT training objective becomes:
\begin{align}
\label{eq:sft}
\mathcal{L}_{\text{SFT}} = -\mathbb{E} \left[ \frac{1}{|M_t^R|} \sum_{i \in M_t^R} \log \mathbb{P}(X^{i}_0 | X^{i}_{t_r}) \right]
\end{align}
where $M_t^R = \{i : R^{i}_{t} = \texttt{[M]}\}$ denotes the set of masked token positions in the response at $t$. This asymmetric masking means the model learns:
\begin{equation}
\mathbb{P}(R_0|P_0, R_t) \quad \text{(response denoising)}
\end{equation}
but never experiences:
\begin{equation}
\mathbb{P}(P_0|P_t, R_0) \quad \text{(prompt denoising)}
\end{equation}
The bidirectional denoising architecture theoretically supports $\mathbb{P}(P_0|P_t, R_0)$, but response-only SFT never provides the necessary training signal.

\section{Additional GSM8K Results}
\label{sec:appendix_gsm8k}

\begin{figure}[h]
\centering
\includegraphics[width=0.5\linewidth]{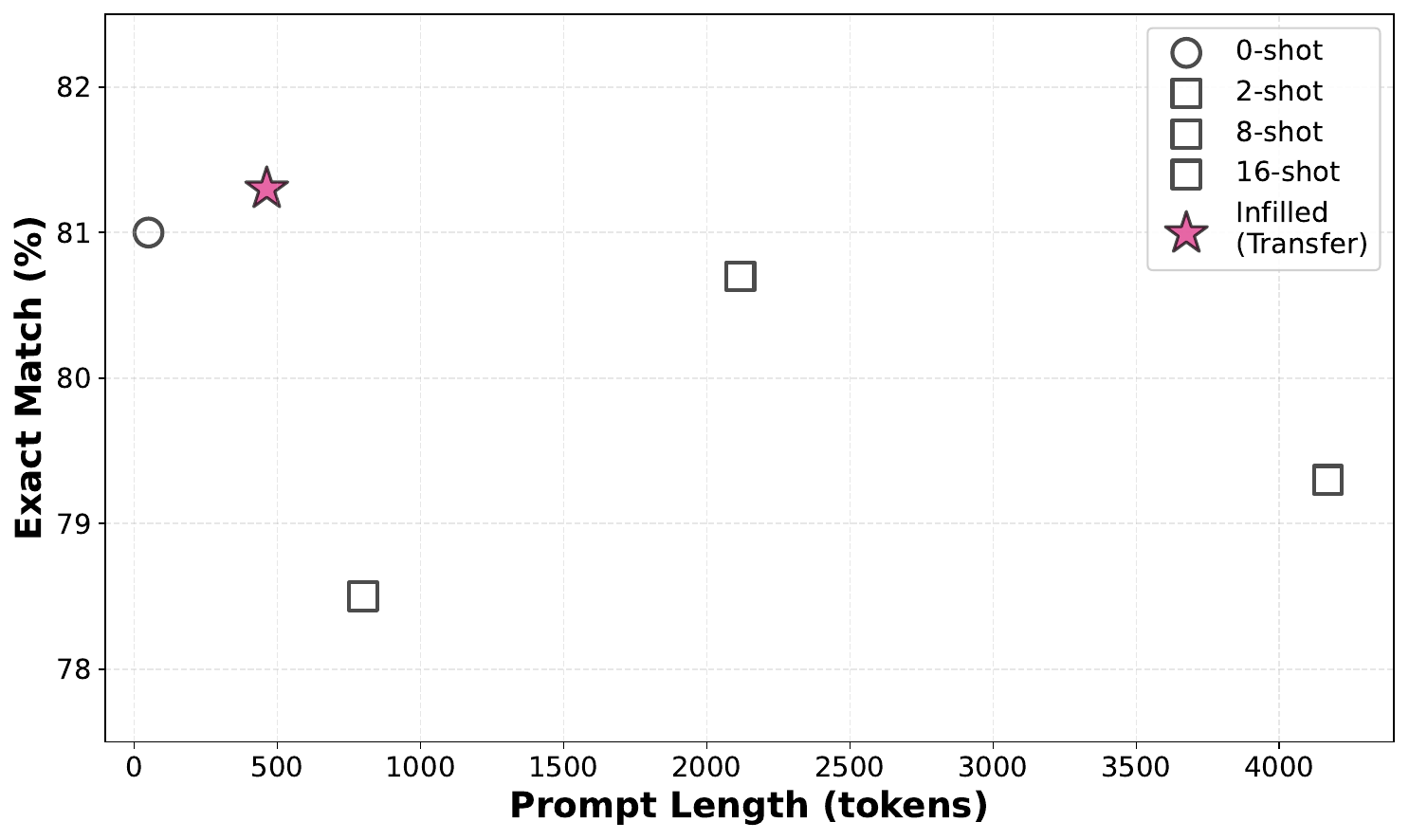}
\caption{GSM8K exact match rate vs. prompt length for Dream public checkpoint. The infilled prompt (red star) achieves the best exact match accuracy (81.3\%) with only 463 tokens on average, outperforming all in-context learning approaches while using significantly fewer tokens than 8-shot (2117 tokens) and 16-shot (4169 tokens) configurations.}
\label{fig:dream_gsm8k}
\end{figure}

Figure~\ref{fig:dream_gsm8k} shows the performance-efficiency trade-off for Dream's public checkpoint across different prompting strategies.

\section{HoVer Results}
\label{sec:appendix_hover}

HoVer~\citep{jiang-etal-2020-hover} requires multi-hop claim verification through iterative retrieval and reasoning, aligning naturally with diffusion models' progressive refinement. The task involves three retrieval hops (initial evidence, refined queries, final verification) followed by classification (SUPPORTED/REFUTED/NOT\_ENOUGH\_INFO). Each hop contains maskable prompt components: query generation instructions, summarization strategies, and evidence aggregation rules.

Table~\ref{tab:hover_fewshot} shows that full-sequence masking (FS) substantially improves label accuracy to $0.550$, outperforming the public checkpoint baseline ($0.435$) by 11.5 percentage points and confirming that full-sequence masking during SFT is essential for unlocking prompt infilling capabilities. Adding response-only refinement (FS+RO) further improves accuracy to $0.604$, suggesting that on this dataset the optional second stage provides additional benefit. Unlike GSM8K's potentially memorized patterns, HoVer's task-specific query generation and evidence aggregation provide a stronger test of genuine prompt infilling capabilities.

\begin{table}[h]
\centering
\small
\begin{tabular}{lcccc}
\toprule
\textbf{Model} & \textbf{Recall} & \textbf{F1} & \textbf{Label Acc}\\
\midrule
LLaDA (Public) & $.560_{.002}$ & $.268_{.001}$ & $.435_{.010}$ \\
LLaDA (FS) & $.556_{.001}$ & $.279_{.001}$ & $.550_{.003}$ \\
LLaDA (RO) & $.533_{.001}$ & $.163_{.001}$ & $.503_{.013}$ \\
LLaDA (FS+RO) & $.547_{.001}$ & $.211_{.000}$ & $.604_{.014}$  \\
\bottomrule
\end{tabular}

\caption{HoVer few-shot evaluation results (mean $\pm$ std over 3 runs). Public: no additional training.}

\label{tab:hover_fewshot}
\end{table}

\raggedbottom

\section{Prompt Templates}
\label{sec:appendix_prompts}

\subsection{Original Feedback Collection Prompt}

\begin{promptbox}[Original Feedback Collection Prompt]
\#\#\#Task Description:\\
An instruction (might include an Input inside it), a response to evaluate, a reference answer that gets a score of 5, and a score rubric representing a evaluation criteria are given.\\
1. Write a detailed feedback that assess the quality of the response strictly based on the given score rubric, not evaluating in general.\\
2. After writing a feedback, write a score that is an integer between 1 and 5. You should refer to the score rubric.\\
3. The output format should look as follows: ``Feedback: (write a feedback for criteria) [RESULT] (an integer number between 1 and 5)''\\
4. Please do not generate any other opening, closing, and explanations.\\

\#\#\#The instruction to evaluate:\\
\{orig\_instruction\}\\

\#\#\#Response to evaluate:\\
\{orig\_response\}\\

\#\#\#Reference Answer (Score 5):

\{orig\_reference\_answer\}\\

\#\#\#Score Rubrics:
[\{orig\_criteria\}]\\
Score 1: \{orig\_score1\_description\}\\
Score 2: \{orig\_score2\_description\}\\
Score 3: \{orig\_score3\_description\}\\
Score 4: \{orig\_score4\_description\}\\
Score 5: \{orig\_score5\_description\}\\

\#\#\#Feedback:

\end{promptbox}

\subsection{Partially Masked Prompt}

\begin{promptbox}[Partially Masked Prompt: Task \& Score]
[Mask] [Mask] [Mask] [Mask] [Mask] [Mask] [Mask] [Mask]...

\#\#\#The instruction to evaluate:\\
\{orig\_instruction\}\\

\#\#\#Response to evaluate:\\
\{orig\_response\}\\

\#\#\#Reference Answer (Score 5):

\{orig\_reference\_answer\}\\

\#\#\#Score Rubrics:
[\{orig\_criteria\}]\\
Score [Mask] [Mask]...\\
Score [Mask] [Mask]...\\
Score [Mask] [Mask]...\\
Score [Mask] [Mask]...\\
Score [Mask] [Mask]...\\

\#\#\#Feedback:

\end{promptbox}

\subsection{Example Infilled Prompt}

The following shows an example prompt after infilling by LLaDA. The task description and score rubric (previously masked) have been infilled from few-shot examples. Note the non-integer score for Score 4 (4.2), which the model inferred from the few-shot examples:

\begin{promptbox}[Infilled Prompt Example]
\#\#\#Task Description:\\
An instruction (might include an Input inside it), a response to evaluate, a reference answer that gets a score of 5, and a score rubric representing a evaluation criteria are given.\\
1. Write a detailed feedback that assess the quality of the response strictly based on the given score rubric, not evaluating in general.\\
2. After writing a feedback, write a score that is a integer number between 1 and 5.\\
3. The output format should look as follows: ``Feedback: (write a feedback for criteria) [RESULT] ( integer number between 1 and 5)''\\

\#\#\#The instruction to evaluate:\\
\{orig\_instruction\}\\

\#\#\#Response to evaluate:\\
\{orig\_response\}\\

\#\#\#Reference Answer (Score 5):

\{orig\_reference\_answer\}\\

\#\#\#Score Rubrics:

[Overall Quality: The overall quality of the summary considering all aspects including coherence, consistency, fluency, and relevance.]\\
Score 1: Score of 1.0\\
Score 2: Score of 1.2\\
Score 3: Score of 1.8\\
Score 4: Score of 4.8\\
Score 5: Score of 5.0\\

\#\#\#Feedback:

\end{promptbox}

\subsection{C-gen Template for SummEval}

This is the C-gen template used for SummEval evaluation:

\begin{promptbox}[C-Gen Prompt Template for SummEval]
Please evaluate the following summary based on the given source text and reference summary.

Source Text:\\
\{source\}

Reference Summary:\\
\{reference\}

Summary to Evaluate:\\
\{system\_output\}

Evaluation Criterion:\\
Overall Quality: The overall quality of the summary considering all aspects including coherence, consistency, fluency, and relevance.

Rate the overall quality of the summary on a scale of 1-5, where 1 is very poor and 5 is excellent.

Provide your evaluation and end with [RESULT] followed by your numerical score (1-5).
\end{promptbox}

\section{Potential Risks and Prompt Leakage}
\label{sec:appendix_risks}

While our approach enables beneficial applications such as prompt optimization and adaptation, we acknowledge potential risks associated with prompt infilling capabilities. One concern is prompt leakage: malicious actors could potentially use prompt infilling to reverse-engineer proprietary prompts from observed model outputs. If a model can infer optimal prompts from responses, the same capability could theoretically be exploited to extract confidential prompt engineering strategies from commercial systems by observing their outputs. This risk is particularly relevant for systems that rely on prompt secrecy as part of their intellectual property or competitive advantage. Future work should investigate defensive mechanisms such as prompt obfuscation techniques, watermarking strategies, or access control policies to mitigate unauthorized prompt extraction while preserving the benefits of prompt infilling for legitimate use cases.

\section{Information About Use of LLMs}
\label{sec:appendix_ai_assistants}

In preparing this manuscript, we used Claude Code for two purposes: (1) revising and editing the manuscript text, including improving clarity, fixing grammatical errors, and refining technical descriptions, and (2) generating and debugging code for running experiments, creating plots, and processing experimental results.

\end{document}